\documentclass[
]{ceurart}

\usepackage{listings}
\usepackage{cleveref}
\begin{document}

\copyrightyear{2022}
\copyrightclause{Copyright for this paper by its authors.
  Use permitted under Creative Commons License Attribution 4.0
  International (CC BY 4.0).}

\conference{xAI'24: 2nd World Conference on eXplainable Artificial Intelligence
  July 17--19, 2024, Valletta, Malta}

\title{Faithful Attention Explainer: Verbalizing Decisions Based on Discriminative Features}

\tnotemark[1]
\tnotetext[1]{You can use this document as the template for preparing your
  publication. We recommend using the latest version of the ceurart style.}

\author[1]{Yao Rong}[%
orcid=0000-0002-6031-3741,
email=yao.rong@tum.de
]
\cormark[1]
\address[1]{Technical University of Munich, Arcisstraße 21, 80333 Munich, Germany}

\author[2]{David Scheerer}[%
email=david.scheerer@student.uni-tuebingen.de
]
\address[2]{University of Tübingen, Sand 14, 72076 Tübingen, Germany}

\author[1]{Enkelejda Kasneci}[%
orcid=0000-0003-3146-4484,
email=enkelejda.kasneci@tum.de,
]


\cortext[1]{Corresponding author.}

\begin{abstract}
  In recent years, model explanation methods have been designed to interpret model decisions faithfully and intuitively so that users can easily understand them. 
  In this paper, we propose a framework, Faithful Attention Explainer (FAE), capable of generating faithful textual explanations regarding the attended-to features. Towards this goal, we deploy an attention module that takes the visual feature maps from the classifier for sentence generation. Furthermore, our method successfully learns the association between features and words, which allows a novel attention enforcement module for attention explanation. Our model achieves promising performance in caption quality metrics and a faithful decision-relevance metric on two datasets (CUB and ACT-X). In addition, we show that FAE can interpret gaze-based human attention, as human gaze indicates the discriminative features that humans use for decision-making, demonstrating the potential of deploying human gaze for advanced human-AI interaction.
\end{abstract}


\maketitle
\section{Introduction}

Explainable AI (XAI) models are being used more, especially in safety-critical applications such as automatic medical diagnosis \cite{pham2021interpreting,tjoa2020survey, rong2022user}. An explanation of a decision should be understandable for humans~\cite{rong2023towards}, and include objects or features that are responsible for that decision made by a model, i.e., faithful to the model decision \cite{Wickramanayake2019FLEX,Hendricks2016GVE, rong2022consistent}. 

In image-based applications, two modalities are typically used in model explanations: visual and textual explanation \cite{Park2018Multimodal}. Several related works in this context \cite{petsiuk2018rise,zhou2016learning,selvaraju2017grad,sundararajan2017axiomatic,shrikumar2017learning} reveal discriminative (salient) areas for the neural network in decision-making by means of saliency maps. Such saliency maps visualize the post-hoc attention of a deep neural network. However, humans often prefer textual justifications of model decisions since they allow for easier access to the understanding of the causality provided by models \cite{Hendricks2016GVE,kim2018textual}. 
In this work, we introduce a novel method, ``Faithful Attention Explainer'' (FAE), which generates faithful textual explanations according to the decision made by the classifier. 
\begin{wrapfigure}{t}{0.5\textwidth}
    \centering
    \includegraphics[width=\linewidth]{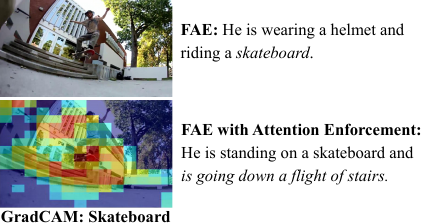}
    \caption{FAE generates faithful explanations (Top). Using attention enforcement, FAE generates a sentence further explaining the attended-to area in GradCAM (Bottom).}
    \vspace{-20pt}
    \label{fig:teaser}
\end{wrapfigure}
As shown by an example in Figure~\ref{fig:teaser}, the explanation of our model includes the object ``skateboard," which is used for the action classification (shown in GradCAM \cite{Selvaraju2016GradCAM}). 
When we give the GradCAM as the extrinsic attention, the model describes more of the area, such as ``standing on a skateboard" and ``going down a flight of stairs." 
Similarly, human attention also conveys the potential to explain our decisions \cite{posner1990attention}. It is visualized in the saliency map style and compared to models' post-hoc attention maps in solving visual question answering and classification tasks \cite{das2017human, rong2021human}. In this context, the language model should also be able to generate a faithful explanation based on human attention. Providing human attention interpretation can help study the human attention mechanism and better integrate it into computer vision applications. To summarize, this work proposes a novel framework, FAE, which generates faithful textual explanations based on attention maps (from models or humans).

\vspace{-5pt}
\section{Related Work}
    


Attention models for generating textual descriptions are known to be highly effective \cite{Liu2020Prophet,Xu2015SAT,chen2017sca,lu2017knowing,you2016image}. For example, \cite{Xu2015SAT} proposes an attention model consisting of linear layers to localize the relevant area in the image for sentence generation. 
However, the attention model grounds the current word to a wrong region since its current hidden state contains only information of past words \cite{Liu2020Prophet}. To solve this problem, \cite{liu2017attention,zhou2019grounded} use extra supervision for correct visual grounding is therefore needed, while \cite{Liu2020Prophet} proposes the Prophet Attention model which takes both future and past words into account and recreates attention weights and thus does not require extra supervision. Inspired by the PA model, we incorporate future words (generated after the current word) to ground the current word in the image in our attention model. 
Generating faithful explanations for classifiers is more than image captioning \cite{Hendricks2016GVE, Wickramanayake2019FLEX} since the generated sentence must rationalize the decision and include discriminative features for the distinctive output class. To generate sentences conditioned on classifiers, previous works \cite{Barratt2017Interp, Hendricks2016GVE, Wickramanayake2019FLEX, Park2018Multimodal,kim2018textual} use features from the corresponding classifier and feed them into an LSTM layer to generate textual explanations. 
However, these explanations may not be faithful to each sample since they are trained to be discriminative on class-level and thus can generate features that are not visible in that image \cite{Wickramanayake2019FLEX}. Going beyond previous work, our framework utilizes an attention module for word grounding directly.
\section{Methodology}

\begin{figure}[]
    \centering
    \includegraphics[width=\linewidth]{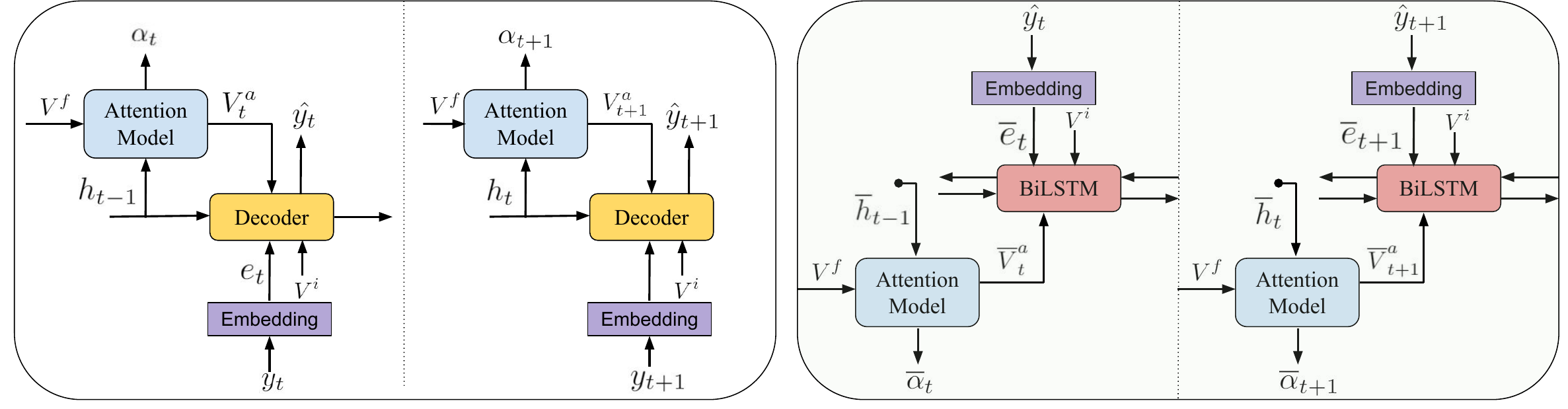}
    \caption{Overview of Faithful Attention Explainer. The encoder is omitted for simplicity but the output features $V^f$ and $V^i$ from the encoder are denoted. The embedding layer is used to transform words into embeddings. \textbf{Left}: the attention model and decoder are illustrated. The attention model produces attention $\alpha$ based on the previous sequence. \textbf{Right}: the attention alignment is used to produce $\hat{\alpha}$ based on the generated sequence $\hat{y}_{t:T}$, which tries to align $\alpha$ with $\hat{\alpha}$.}
    \label{fig:FAE model}
    \vspace{-15pt}
\end{figure}

Our FAE generates textual explanations for image classifiers, i.e., FAE verbalizes classification decisions by creating sentences containing words related to image regions that have been important to the decision of the classifier.
In this section, we explain the details of each module in FAE and introduce the Attention Enforcement algorithm in detail.

Our network approach follows an Encoder-Decoder framework. 
The goal of FAE is to take the image $x \in \mathbb{R}^{H\times W \times C} $ and to predict the class label as well as to create a textual explanation $\mathbf{\hat{y}}$ as a sequence of 1-of-N words:
\begin{equation}
    \mathbf{\hat{y}} = \{ \hat{y_1}, \hat{y_2}, ..., \hat{y_T} \}, \indent \hat{y_t} \in \mathbb{R}^N
\end{equation}
where T denotes the length of the output and $y_t$ is the predicted word at $t$ step.
FAE exploits the class-discriminative feature vector $V \in \mathbb{R}^{h\times w \times c}$ from the classifier (also used as the encoder $\Sigma(\cdot)$). $\Sigma$ is a deep convolutional neural network and can extract several visual feature vectors $V$ from different layers of the input image $x$. Taking ResNet101 as an example, $V^f$ is the feature map after the last residual block, while $V^i$ can be a set of feature maps taken from the final layer of the first, second, and third blocks. 

For each step $t$, the attention model $f_{Att}(\cdot)$ computes attention maps $\alpha_t$ based on the decoder's (an LSTM model) hidden state $h_{t-1}$ and the feature vector from the encoder. The output of the attention module $V_t^a$ is given to the decoder and guides it towards important areas relevant for the explanation:
the attention-weighted average of focus features $V^a_t$ is:
\begin{equation}
        V^a_{t} = \frac{1}{K} \sum_{j=1}^K \alpha_{t,j} V^f_j.
\end{equation}

\Cref{fig:FAE model} (Left) illustrates this architecture that contains the attention module for generating textual explanations.
We follow the method proposed in~\cite{Xu2015SAT} to build and train this model. As the attention model computes weights based on the previous hidden state of the LSTM, which is generated using the previous input word. As a result, the attention weights are also based on the previous word. 
To tackle this challenge, we introduce a module called attention alignment. Inside the module, we make use of future knowledge (words) to adjust the attention map for the current word. 
To do so, a Bidirectional LSTM (BiLSTM)\cite{schuster1997bidirectional} is employed to encode the generated sequence. The attention model described in the last section is used to regenerate new attention weights $\overline{\alpha}_t$ based on the hidden state $\overline{h}_{t-1}$ from the BiLSTM.
Specifically, we get $\overline{h}_{t-1}$ by concatenating the hidden states from forward and backward paths (and halving the dimension). \Cref{fig:FAE model} (Right) illustrates the attention alignment.
\begin{equation}
    \begin{split}
        \overline{\alpha}_{t} = f_{Att}(\overline{h}_{t-1}, V^f)\\
        \overline{V}^a_t = \frac{1}{K} \sum_{j=1}^K \overline{\alpha}_t V^f_j
    \end{split}
\end{equation}
As a regularization to the training loss, we use the L1 norm between the newly grounded attention weights $\overline{\alpha}$ and the ones generated by the attention model $\alpha$:
\begin{equation}
  \mathcal{L}_{\alpha}(\theta) = \sum^T_{t=1} ||\alpha_t - \overline{\alpha}_t ||
\end{equation}

Moreover, the learned attention can be given by users, i.e., by replacing attention weights by other attention maps $\epsilon$, e.g., GradCAM or human gaze, during inference. We refer this as Attention Enforcement (AE). Concretely, we generate the focus feature $V^{\epsilon}$:
\begin{equation}
    \begin{split}
        {V^{\epsilon}} = \frac{1}{K} \sum_{j=1}^K \text{Softmax}(\epsilon) V^f_j
    \end{split}
\end{equation}
\section{Experiments}
\paragraph{Metrics.} To evaluate and compare our model with other works, we use the following metrics: BLEU-4, ROGUE-L, METEOR, CIDer. These metrics measure the similarity between generated sentences and their ground-truth. However, they only indicate the sentence quality on a linguistic level but have no insights into the faithfulness of generated explanations. Therefore, we measure the Faithful Explanation Rate (\textit{FER}) in generated explanations compared to ground-truth sentences, inspired by \cite{Wickramanayake2019FLEX}. Specifically, for an image $x$, discriminative visual regions used in the model's decision are found out with the help of GradCAM \cite{Selvaraju2016GradCAM}. Using the part annotations, the decision-related part/object $y_o$ can be identified (the part that is closest to the maximum value in GradCAM). Noun-phrases of that part in all ground-truth sentences are extracted to form a set $\{\mathbf{g}_1, \mathbf{g}_2,...,\mathbf{g}_M\}$ where $\mathbf{g}_i$ denotes for a noun-phrase. For the generated sequence $\mathbf{\hat{y}}$, we detect whether the $y_o$ is in $\mathbf{\hat{y}}$, if not, the hit rate is 0. If yes, we detect the corresponding noun-phrase $\mathbf{\hat{g}}$. Then we compare the word hit rate of $\mathbf{\hat{g}}$ with all possible $\mathbf{g}_i$ and use the best one for the FER score.




\paragraph{Datasets.} We use two datasets for our experiments: the CUB-200-2011 dataset (CUB) and Action Explanation Dataset (ACT-X). CUB contains 11.788 images of birds distributed across 200 species \cite{WahCUB}. Each image has ten explanations of the visual appearance collected by \cite{Reed2016CUBCaps}. ACT-X \cite{Park2018Multimodal} has 397 classes of activities and in total 18030 images selected from \cite{andriluka14MPII}. For each image, three explanations are provided. We follow the provided train and test splits on both datasets. When evaluating the FER score, we use the part annotations on CUB and object-level annotations on ACT-X. The object-level annotation on ACT-X denoted as MPII-ANO, only contains a few images in ACT-X (150 images with 600 object classes) provided by \cite{Wickramanayake2019FLEX}.

\subsection{Quantitative Results}
\begin{table}[t]
        \resizebox{.6\linewidth}{!}{
        \renewcommand{\arraystretch}{1.1}
      \begin{tabular}{cccccc}
        \toprule
        Dataset & Method & Backbone  & BLEU-4 & METEOR & CIDer \\ 
        \midrule
    \multirow{6}{*}{CUB}      
    
    & GVE \cite{Hendricks2016GVE} & VGG   &- &29.20 &56.70 \\ 
        & InterpNET \cite{Barratt2017Interp}  & VGG    &\textbf{62.30} &37.90& \textbf{82.10} \\ 
    
        &  SAT & ResNet-101    &57.14 &36.71 &61.80 \\
        &  FAE (Ours)   &ResNet-50     &57.94  & 36.33 &55.98  \\  
        &  FAE (Ours)  & ResNet-101      & {60.19} &\textbf{38.13} & {66.36} \\  
      \midrule
    \multirow{6}{*}{ACT-X} & GVE \cite{Hendricks2016GVE} & VGG & 12.90 & 15.90 & 12.40 \\
            & PJ-X \cite{Park2018Multimodal} & ResNet-152 & 24.50 & 21.50 & {58.70} \\
            & SAT \cite{Xu2015SAT} & ResNet-101 & 25.63 & {24.53} & 50.39 \\
            & FAE (Ours)  & ResNet-50 & {26.66}  & 24.37 & 57.19  \\
            & FAE (Ours) & ResNet-101 & \textbf{27.06} &\textbf{25.33} &\textbf{66.17} \\
        \bottomrule
      \end{tabular}
     }  
\hfill
    \resizebox{.38\linewidth}{!}{
        \begin{tabular}{c|c|c}
        \hline
        Method & CUB   & MPII-ANO \\ \hline
        SAT \cite{Xu2015SAT} &37.43 &26.32 \\ \hline
        FAE (Ours)   & 39.42 &     28.40 \\ \hline \hline
        SAT-AE \cite{Xu2015SAT}& 38.54 & 26.84    \\ \hline
        FAE-AE (Ours) & \textbf{44.33} & \textbf{29.76}   \\ \hline
        \end{tabular}
    }
\caption{\textbf{Left:} Comparison with other methods on CUB and ACT-X in standard sentence quality metrics. \textbf{Right:} FER score on CUB and MPII-ANO. The first block contains methods without Attention Enforcement (AE); the second block with using AE. ResNet101 is used as the backbone for all models. }
\vspace{-15pt}
 \label{tab:comp_sota}
\end{table}


We first compare our model with other state-of-the-art approaches in the linguistic quality of generated explanations. In \cref{tab:comp_sota} (Left), we compare our FAE using two backbones with
InterpNET \cite{Barratt2017Interp}, Generating Visual Explanations (GVE) \cite{Hendricks2016GVE}, and Pointing and Justification Explanation (PJ-X) model \cite{Park2018Multimodal}. 
On CUB, our model (using ResNet101 backbone) outperforms GVE, e.g., in the metric CIDer, our model achieves 66.36 while GVE achieves 56.70. Compared to InterpNET, however, our model surpasses only in METEOR. The possible reason is that InterpNET deploys richer features (8192-dim compact bilinear features), two extra hidden layers, and two stacked LSTM layers, which introduces more computational costs and makes the results hard to reproduce. Results on ACT-X are shown in the second block. Our model (ResNet101) achieves higher scores in all three metrics than other methods. 

Besides the linguistic quality, FER score are shown in \cref{tab:comp_sota} (Right). We compare our framework with SAT since Attention Enforcement (AE) can also be applied to it. For a fair comparison, we evaluate both under the same settings. In the first block, where no AE is used, FAE achieves the best performance: 39.42 on CUB and 28.40 on MPII-ANO, which validates that our FAE is advanced in faithful explanation generation. When using GradCAM attention enforcement, SAT and FAE both improve the FER scores, while FAE surpasses SAT on both datasets. The improvement of using AE in both models validates the generalization of AE. 

\subsection{Qualitative Results}
\label{sec:qualitative}

\begin{figure*}[t]
    \includegraphics[width=\linewidth]{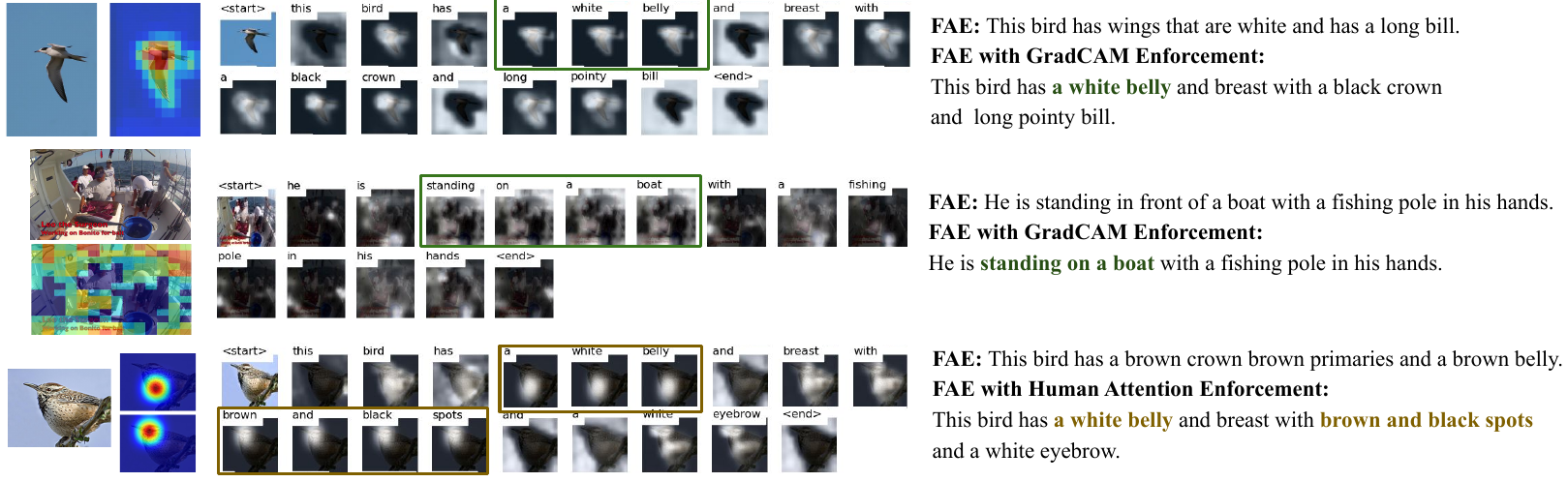}
    \caption{Illustration of using attention enforcement on CUB and MPII-ANO. \textbf{Left:} Images and extrinsic saliency maps are shown. \textbf{Middle:} Frames denote the step where enforcement is activated. \textbf{Right:} Sentences generated by FAE with and without attention enforcement. The top two examples use GradCAM from the classifier as extrinsic attention maps, while the bottom one uses human gaze maps.}
    \vspace{-15pt}
    \label{fig:fae ae}
\end{figure*}

We give GradCAM maps as extrinsic attention maps to guide the model FAE with AE to focus on the area highlighted in the attention map. Two generated sentence examples are illustrated in \Cref{fig:fae ae}. After applying the enforcement in the first example, the explanation incorporates the part ``a white belly", which is missing before. Nevertheless, when enforcement on the MPII-ANO dataset, the effects are others. Since the GradCAM highlights a lot of area on the boat and in the background (on the sea), the sentence after the enforcement describes the relation between objects correctly: the man is standing ``on the boat" instead of ``in front of a boat". The results show that our FAE can provide explanations that are faithful and human-understandable to not only intrinsic but also extrinsic attention maps.

Additionally, we try a different source of extrinsic attention for AE: Human Attention (HA). We evaluate our HA-enforcement on the CUB test set and use the HA map provided in CUB Gaze-based Human Attention (CUB-GHA) \cite{rong2021human}. This dataset is built by tracking the eye fixations of humans while presenting them of a bird to focus on distinctive features for that species. For each image, there are always multiple attention maps and each attention map represents an eye fixation. In \Cref{fig:fae ae}, the bottom example shows the HA attention maps. When we deploy our AE on using HA as extrinsic attention information, the sentence describes the two areas: ``a white belly" in the first fixation area and ``breast with brown and black spots" in the second attention area. 
This setting confirms that our method can produce accurate textual explanations focusing on user attention, demonstrating the generalizability of our proposed framework.

\section{Discussion}
\begin{figure}[t]
    \centering
\includegraphics[width=.9\linewidth]{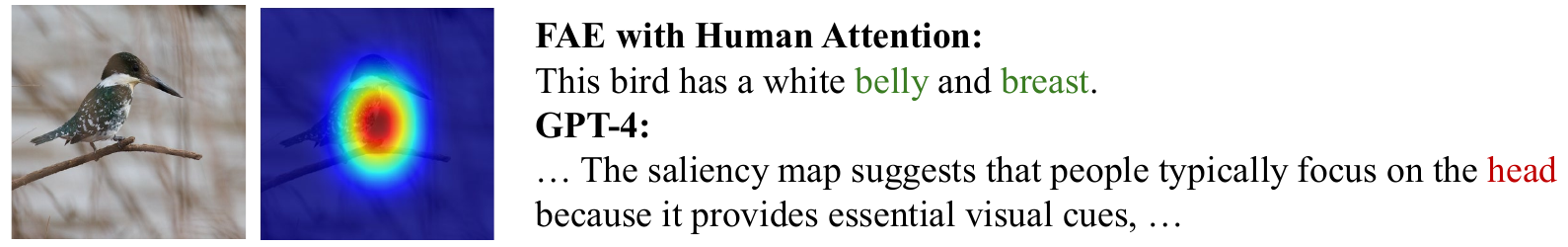}
    \caption{Comparison of our method and GPT-4 in generating textual explanations.}
    \vspace{-15pt}
    \label{fig:gpt}
\end{figure}

Large Language Models (LLMs), such as the GPT series, have demonstrated their sophisticated abilities in understanding and generating explanations. Recent advancements enable these models to analyze multimodal data. For example, the GPT-4 model can create textual explanations from an input image. To evaluate its effectiveness, we tested the GPT-4 model with two types of images: an original image and a saliency map highlighting human attention, as illustrated in \Cref{fig:gpt}. The GPT-4 successfully generated an analysis of the areas most salient to human gaze. However, we observe the problem in the generated textual explanations: the model fails to correctly identify the area where the user focused. For example, it mistook the belly/breast area as the head. These mistakes rather demonstrate a common weakness in the model: hallucination.
To harvest the power of language models, we consider for future work fine-tuning a smaller general language model to generate textual explanations based on the areas of gaze attention of users. This approach can enhance the possibilities for intuitive and direct interaction between humans and AI systems through gaze-based communication.

\section{Conclusion}
In this paper, we propose a novel framework FAE that can generate decision explanations faithful to intrinsic attention, i.e., generated by an attention model based on visual features from the classifier. Our results on the CUB and ACT-X datasets validate and confirm the high faithfulness and quality in explanations provided by FAE. Moreover, we extend FAE by using Attention Enforcement and can thus interpret extrinsic attention e.g., human attention. For future work, our method expands opportunities for natural and straightforward communication between humans and AI systems via gaze-driven interactions. 

\bibliography{sample-ceur}

\begin{thebibliography}{30}
\expandafter\ifx\csname natexlab\endcsname\relax\def\natexlab#1{#1}\fi
\providecommand{\url}[1]{\texttt{#1}}
\providecommand{\href}[2]{#2}
\providecommand{\path}[1]{#1}
\providecommand{\DOIprefix}{doi:}
\providecommand{\ArXivprefix}{arXiv:}
\providecommand{\URLprefix}{URL: }
\providecommand{\Pubmedprefix}{pmid:}
\providecommand{\doi}[1]{\href{http://dx.doi.org/#1}{\path{#1}}}
\providecommand{\Pubmed}[1]{\href{pmid:#1}{\path{#1}}}
\providecommand{\bibinfo}[2]{#2}
\ifx\xfnm\relax \def\xfnm[#1]{\unskip,\space#1}\fi
\bibitem[{Pham et~al.(2021)Pham, Le, Tran, Ngo, and
  Nguyen}]{pham2021interpreting}
\bibinfo{author}{H.~H. Pham}, \bibinfo{author}{T.~T. Le},
  \bibinfo{author}{D.~Q. Tran}, \bibinfo{author}{D.~T. Ngo},
  \bibinfo{author}{H.~Q. Nguyen},
\newblock \bibinfo{title}{Interpreting chest x-rays via cnns that exploit
  hierarchical disease dependencies and uncertainty labels},
\newblock \bibinfo{journal}{Neurocomputing}  (\bibinfo{year}{2021}).
\bibitem[{Tjoa and Guan(2019)}]{tjoa2020survey}
\bibinfo{author}{E.~Tjoa}, \bibinfo{author}{C.~Guan},
\newblock \bibinfo{title}{A survey on explainable artificial intelligence
  {(XAI):} towards medical {XAI}},
\newblock \bibinfo{journal}{CoRR}  (\bibinfo{year}{2019}). \URLprefix
  \url{http://arxiv.org/abs/1907.07374}.
  \href{http://arxiv.org/abs/1907.07374}{{\tt arXiv:1907.07374}}.
\bibitem[{Rong et~al.(2022)Rong, Castner, Bozkir, and Kasneci}]{rong2022user}
\bibinfo{author}{Y.~Rong}, \bibinfo{author}{N.~Castner},
  \bibinfo{author}{E.~Bozkir}, \bibinfo{author}{E.~Kasneci},
\newblock \bibinfo{title}{User trust on an explainable ai-based medical
  diagnosis support system},
\newblock \bibinfo{journal}{arXiv preprint arXiv:2204.12230}
  (\bibinfo{year}{2022}).
\bibitem[{Rong et~al.(2023)Rong, Leemann, Nguyen, Fiedler, Qian, Unhelkar,
  Seidel, Kasneci, and Kasneci}]{rong2023towards}
\bibinfo{author}{Y.~Rong}, \bibinfo{author}{T.~Leemann}, \bibinfo{author}{T.-T.
  Nguyen}, \bibinfo{author}{L.~Fiedler}, \bibinfo{author}{P.~Qian},
  \bibinfo{author}{V.~Unhelkar}, \bibinfo{author}{T.~Seidel},
  \bibinfo{author}{G.~Kasneci}, \bibinfo{author}{E.~Kasneci},
\newblock \bibinfo{title}{Towards human-centered explainable ai: A survey of
  user studies for model explanations},
\newblock \bibinfo{journal}{IEEE Transactions on Pattern Analysis and Machine
  Intelligence}  (\bibinfo{year}{2023}).
\bibitem[{Wickramanayake et~al.(2019)Wickramanayake, Hsu, and
  Lee}]{Wickramanayake2019FLEX}
\bibinfo{author}{S.~Wickramanayake}, \bibinfo{author}{W.~Hsu},
  \bibinfo{author}{M.~Lee},
\newblock \bibinfo{title}{Flex: Faithful linguistic explanations for neural net
  based model decisions},
\newblock in: \bibinfo{booktitle}{AAAI}, \bibinfo{year}{2019}.
\bibitem[{Hendricks et~al.(2016)Hendricks, Akata, Rohrbach, Donahue, Schiele,
  and Darrell}]{Hendricks2016GVE}
\bibinfo{author}{L.~A. Hendricks}, \bibinfo{author}{Z.~Akata},
  \bibinfo{author}{M.~Rohrbach}, \bibinfo{author}{J.~Donahue},
  \bibinfo{author}{B.~Schiele}, \bibinfo{author}{T.~Darrell},
\newblock \bibinfo{title}{Generating visual explanations},
\newblock \bibinfo{journal}{CoRR}  (\bibinfo{year}{2016}). \URLprefix
  \url{http://arxiv.org/abs/1603.08507}.
  \href{http://arxiv.org/abs/1603.08507}{{\tt arXiv:1603.08507}}.
\bibitem[{Rong et~al.(2022)Rong, Leemann, Borisov, Kasneci, and
  Kasneci}]{rong2022consistent}
\bibinfo{author}{Y.~Rong}, \bibinfo{author}{T.~Leemann},
  \bibinfo{author}{V.~Borisov}, \bibinfo{author}{G.~Kasneci},
  \bibinfo{author}{E.~Kasneci},
\newblock \bibinfo{title}{A consistent and efficient evaluation strategy for
  attribution methods},
\newblock in: \bibinfo{booktitle}{International Conference on Machine
  Learning}, \bibinfo{organization}{PMLR}, \bibinfo{year}{2022}, pp.
  \bibinfo{pages}{18770--18795}.
\bibitem[{Park et~al.(2018)Park, Hendricks, Akata, Rohrbach, Schiele, Darrell,
  and Rohrbach}]{Park2018Multimodal}
\bibinfo{author}{D.~H. Park}, \bibinfo{author}{L.~A. Hendricks},
  \bibinfo{author}{Z.~Akata}, \bibinfo{author}{A.~Rohrbach},
  \bibinfo{author}{B.~Schiele}, \bibinfo{author}{T.~Darrell},
  \bibinfo{author}{M.~Rohrbach},
\newblock \bibinfo{title}{Multimodal explanations: Justifying decisions and
  pointing to the evidence},
\newblock \bibinfo{journal}{CoRR}  (\bibinfo{year}{2018}). \URLprefix
  \url{http://arxiv.org/abs/1802.08129}.
  \href{http://arxiv.org/abs/1802.08129}{{\tt arXiv:1802.08129}}.
\bibitem[{Petsiuk et~al.(2018)Petsiuk, Das, and Saenko}]{petsiuk2018rise}
\bibinfo{author}{V.~Petsiuk}, \bibinfo{author}{A.~Das},
  \bibinfo{author}{K.~Saenko},
\newblock \bibinfo{title}{Rise: Randomized input sampling for explanation of
  black-box models},
\newblock \bibinfo{journal}{BMVC}  (\bibinfo{year}{2018}).
\bibitem[{Zhou et~al.(2016)Zhou, Khosla, Lapedriza, Oliva, and
  Torralba}]{zhou2016learning}
\bibinfo{author}{B.~Zhou}, \bibinfo{author}{A.~Khosla},
  \bibinfo{author}{A.~Lapedriza}, \bibinfo{author}{A.~Oliva},
  \bibinfo{author}{A.~Torralba},
\newblock \bibinfo{title}{Learning deep features for discriminative
  localization},
\newblock in: \bibinfo{booktitle}{CVPR}, \bibinfo{year}{2016}.
\bibitem[{Selvaraju et~al.(2017)Selvaraju, Cogswell, Das, Vedantam, Parikh, and
  Batra}]{selvaraju2017grad}
\bibinfo{author}{R.~R. Selvaraju}, \bibinfo{author}{M.~Cogswell},
  \bibinfo{author}{A.~Das}, \bibinfo{author}{R.~Vedantam},
  \bibinfo{author}{D.~Parikh}, \bibinfo{author}{D.~Batra},
\newblock \bibinfo{title}{Grad-cam: Visual explanations from deep networks via
  gradient-based localization},
\newblock in: \bibinfo{booktitle}{ICCV}, \bibinfo{year}{2017}.
\bibitem[{Sundararajan et~al.(2017)Sundararajan, Taly, and
  Yan}]{sundararajan2017axiomatic}
\bibinfo{author}{M.~Sundararajan}, \bibinfo{author}{A.~Taly},
  \bibinfo{author}{Q.~Yan},
\newblock \bibinfo{title}{Axiomatic attribution for deep networks},
\newblock in: \bibinfo{booktitle}{ICML}, \bibinfo{year}{2017}.
\bibitem[{Shrikumar et~al.(2017)Shrikumar, Greenside, and
  Kundaje}]{shrikumar2017learning}
\bibinfo{author}{A.~Shrikumar}, \bibinfo{author}{P.~Greenside},
  \bibinfo{author}{A.~Kundaje},
\newblock \bibinfo{title}{Learning important features through propagating
  activation differences},
\newblock in: \bibinfo{booktitle}{ICML}, \bibinfo{year}{2017}.
\bibitem[{Kim et~al.(2018)Kim, Rohrbach, Darrell, Canny, and
  Akata}]{kim2018textual}
\bibinfo{author}{J.~Kim}, \bibinfo{author}{A.~Rohrbach},
  \bibinfo{author}{T.~Darrell}, \bibinfo{author}{J.~F. Canny},
  \bibinfo{author}{Z.~Akata},
\newblock \bibinfo{title}{Textual explanations for self-driving vehicles},
\newblock \bibinfo{journal}{CoRR}  (\bibinfo{year}{2018}). \URLprefix
  \url{http://arxiv.org/abs/1807.11546}.
  \href{http://arxiv.org/abs/1807.11546}{{\tt arXiv:1807.11546}}.
\bibitem[{Selvaraju et~al.(2016)Selvaraju, Das, Vedantam, Cogswell, Parikh, and
  Batra}]{Selvaraju2016GradCAM}
\bibinfo{author}{R.~R. Selvaraju}, \bibinfo{author}{A.~Das},
  \bibinfo{author}{R.~Vedantam}, \bibinfo{author}{M.~Cogswell},
  \bibinfo{author}{D.~Parikh}, \bibinfo{author}{D.~Batra},
\newblock \bibinfo{title}{Grad-cam: Why did you say that? visual explanations
  from deep networks via gradient-based localization},
\newblock \bibinfo{journal}{CoRR}  (\bibinfo{year}{2016}).
\bibitem[{Posner and Petersen(1990)}]{posner1990attention}
\bibinfo{author}{M.~I. Posner}, \bibinfo{author}{S.~E. Petersen},
\newblock \bibinfo{title}{The attention system of the human brain},
\newblock \bibinfo{journal}{Annual review of neuroscience}
  (\bibinfo{year}{1990}).
\bibitem[{Das et~al.(2017)Das, Agrawal, Zitnick, Parikh, and
  Batra}]{das2017human}
\bibinfo{author}{A.~Das}, \bibinfo{author}{H.~Agrawal},
  \bibinfo{author}{L.~Zitnick}, \bibinfo{author}{D.~Parikh},
  \bibinfo{author}{D.~Batra},
\newblock \bibinfo{title}{Human attention in visual question answering: Do
  humans and deep networks look at the same regions?},
\newblock \bibinfo{journal}{Computer Vision and Image Understanding}
  (\bibinfo{year}{2017}).
\bibitem[{Rong et~al.(2021)Rong, Xu, Akata, and Kasneci}]{rong2021human}
\bibinfo{author}{Y.~Rong}, \bibinfo{author}{W.~Xu}, \bibinfo{author}{Z.~Akata},
  \bibinfo{author}{E.~Kasneci},
\newblock \bibinfo{title}{Human attention in fine-grained classification},
\newblock \bibinfo{journal}{arXiv preprint arXiv:2111.01628}
  (\bibinfo{year}{2021}).
\bibitem[{Liu et~al.(2020)Liu, Ren, Wu, Ge, Fan, Zou, and Sun}]{Liu2020Prophet}
\bibinfo{author}{F.~Liu}, \bibinfo{author}{X.~Ren}, \bibinfo{author}{X.~Wu},
  \bibinfo{author}{S.~Ge}, \bibinfo{author}{W.~Fan}, \bibinfo{author}{Y.~Zou},
  \bibinfo{author}{X.~Sun},
\newblock \bibinfo{title}{Prophet attention: Predicting attention with future
  attention},
\newblock in: \bibinfo{booktitle}{NeurIPS}, \bibinfo{year}{2020}. \URLprefix
  \url{https://proceedings.neurips.cc/paper/2020/file/13fe9d84310e77f13a6d184dbf1232f3-Paper.pdf}.
\bibitem[{Xu et~al.(2015)Xu, Ba, Kiros, Cho, Courville, Salakhutdinov, Zemel,
  and Bengio}]{Xu2015SAT}
\bibinfo{author}{K.~Xu}, \bibinfo{author}{J.~Ba}, \bibinfo{author}{R.~Kiros},
  \bibinfo{author}{K.~Cho}, \bibinfo{author}{A.~C. Courville},
  \bibinfo{author}{R.~Salakhutdinov}, \bibinfo{author}{R.~S. Zemel},
  \bibinfo{author}{Y.~Bengio},
\newblock \bibinfo{title}{Show, attend and tell: Neural image caption
  generation with visual attention},
\newblock \bibinfo{journal}{CoRR}  (\bibinfo{year}{2015}). \URLprefix
  \url{http://arxiv.org/abs/1502.03044}.
  \href{http://arxiv.org/abs/1502.03044}{{\tt arXiv:1502.03044}}.
\bibitem[{Chen et~al.(2017)Chen, Zhang, Xiao, Nie, Shao, Liu, and
  Chua}]{chen2017sca}
\bibinfo{author}{L.~Chen}, \bibinfo{author}{H.~Zhang},
  \bibinfo{author}{J.~Xiao}, \bibinfo{author}{L.~Nie},
  \bibinfo{author}{J.~Shao}, \bibinfo{author}{W.~Liu}, \bibinfo{author}{T.-S.
  Chua},
\newblock \bibinfo{title}{Sca-cnn: Spatial and channel-wise attention in
  convolutional networks for image captioning},
\newblock in: \bibinfo{booktitle}{CVPR}, \bibinfo{year}{2017}.
\bibitem[{Lu et~al.(2017)Lu, Xiong, Parikh, and Socher}]{lu2017knowing}
\bibinfo{author}{J.~Lu}, \bibinfo{author}{C.~Xiong},
  \bibinfo{author}{D.~Parikh}, \bibinfo{author}{R.~Socher},
\newblock \bibinfo{title}{Knowing when to look: Adaptive attention via a visual
  sentinel for image captioning},
\newblock in: \bibinfo{booktitle}{CVPR}, \bibinfo{year}{2017}.
\bibitem[{You et~al.(2016)You, Jin, Wang, Fang, and Luo}]{you2016image}
\bibinfo{author}{Q.~You}, \bibinfo{author}{H.~Jin}, \bibinfo{author}{Z.~Wang},
  \bibinfo{author}{C.~Fang}, \bibinfo{author}{J.~Luo},
\newblock \bibinfo{title}{Image captioning with semantic attention},
\newblock in: \bibinfo{booktitle}{CVPR}, \bibinfo{year}{2016}.
\bibitem[{Liu et~al.(2017)Liu, Mao, Sha, and Yuille}]{liu2017attention}
\bibinfo{author}{C.~Liu}, \bibinfo{author}{J.~Mao}, \bibinfo{author}{F.~Sha},
  \bibinfo{author}{A.~Yuille},
\newblock \bibinfo{title}{Attention correctness in neural image captioning},
\newblock in: \bibinfo{booktitle}{AAAI}, \bibinfo{year}{2017}.
\bibitem[{Zhou et~al.(2019)Zhou, Kalantidis, Chen, Corso, and
  Rohrbach}]{zhou2019grounded}
\bibinfo{author}{L.~Zhou}, \bibinfo{author}{Y.~Kalantidis},
  \bibinfo{author}{X.~Chen}, \bibinfo{author}{J.~J. Corso},
  \bibinfo{author}{M.~Rohrbach},
\newblock \bibinfo{title}{Grounded video description},
\newblock in: \bibinfo{booktitle}{CVPR}, \bibinfo{year}{2019}.
\bibitem[{Barratt(2017)}]{Barratt2017Interp}
\bibinfo{author}{S.~Barratt},
\newblock \bibinfo{title}{Interpnet: Neural introspection for interpretable
  deep learning},
\newblock \bibinfo{journal}{ArXiv}  (\bibinfo{year}{2017}).
\bibitem[{Schuster and Paliwal(1997)}]{schuster1997bidirectional}
\bibinfo{author}{M.~Schuster}, \bibinfo{author}{K.~K. Paliwal},
\newblock \bibinfo{title}{Bidirectional recurrent neural networks},
\newblock \bibinfo{journal}{IEEE transactions on Signal Processing}
  (\bibinfo{year}{1997}).
\bibitem[{Wah et~al.(2011)Wah, Branson, Welinder, Perona, and
  Belongie}]{WahCUB}
\bibinfo{author}{C.~Wah}, \bibinfo{author}{S.~Branson},
  \bibinfo{author}{P.~Welinder}, \bibinfo{author}{P.~Perona},
  \bibinfo{author}{S.~Belongie}, \bibinfo{title}{{The Caltech-UCSD
  Birds-200-2011 Dataset}}, \bibinfo{type}{Technical Report}, California
  Institute of Technology, \bibinfo{year}{2011}.
\bibitem[{Reed et~al.(2016)Reed, Akata, Schiele, and Lee}]{Reed2016CUBCaps}
\bibinfo{author}{S.~E. Reed}, \bibinfo{author}{Z.~Akata},
  \bibinfo{author}{B.~Schiele}, \bibinfo{author}{H.~Lee},
\newblock \bibinfo{title}{Learning deep representations of fine-grained visual
  descriptions},
\newblock \bibinfo{journal}{CoRR}  (\bibinfo{year}{2016}). \URLprefix
  \url{http://arxiv.org/abs/1605.05395}.
  \href{http://arxiv.org/abs/1605.05395}{{\tt arXiv:1605.05395}}.
\bibitem[{Andriluka et~al.(2014)Andriluka, Pishchulin, Gehler, and
  Schiele}]{andriluka14MPII}
\bibinfo{author}{M.~Andriluka}, \bibinfo{author}{L.~Pishchulin},
  \bibinfo{author}{P.~Gehler}, \bibinfo{author}{B.~Schiele},
\newblock \bibinfo{title}{2d human pose estimation: New benchmark and state of
  the art analysis},
\newblock in: \bibinfo{booktitle}{IEEE Conference on Computer Vision and
  Pattern Recognition (CVPR)}, \bibinfo{year}{2014}.

\end{thebibliography}




\end{document}